\documentclass[conference]{IEEEtran}
\IEEEoverridecommandlockouts
\usepackage{cite}
\usepackage{amsmath,amssymb,amsfonts}
\usepackage{float}
\usepackage{algorithmic}
\usepackage{makecell}
\usepackage{graphicx}
\usepackage{textcomp}
\usepackage{xcolor}
\def\BibTeX{{\rm B\kern-.05em{\sc i\kern-.025em b}\kern-.08em
    T\kern-.1667em\lower.7ex\hbox{E}\kern-.125emX}}
\begin{document}

\title{Emissions and Performance Trade-off Between Small and Large Language Models}

\author{
  \begin{tabular}{cc}
    \begin{tabular}[t]{c} 
       Anandita Garg \\
      {Harish and Bina Shah School} \\
      {of Computer Science and Artificial Intelligence} \\
      {Plaksha University} \\
      {Mohali, India} \\
      {anandita.garg.ug23@plaksha.edu.in}
    \end{tabular}
    &  
    \begin{tabular}[t]{c} 
       Uma Gaba \\
      {Harish and Bina Shah School} \\
      {of Computer Science and Artificial Intelligence} \\
      {Plaksha University} \\
      {Mohali, India} \\
      {uma.gaba.ug23@plaksha.edu.in}
    \end{tabular}
    \\ 
    
    \\[2ex] 

    \begin{tabular}[t]{c} 
       Deepan Muthirayan \\
      {Harish and Bina Shah School} \\
      {of Computer Science and Artificial Intelligence} \\
      {Plaksha University} \\
      {Mohali, India} \\
      {deepan.muthirayan@plaksha.edu.in}
    \end{tabular}
    & 
    \begin{tabular}[t]{c} 
       Anish Roy Chowdhury \\
      {Harish and Bina Shah School} \\
      {of Computer Science and Artificial Intelligence} \\
      {Plaksha University} \\
      {Mohali, India} \\
      {anish.chowdhury@plaksha.edu.in}
    \end{tabular}
  \end{tabular}
}
\maketitle

\begin{abstract}
The advent of Large Language Models (LLMs) has
raised concerns about their enormous carbon footprint, starting with energy-intensive training and continuing through repeated
inference. This study investigates the potential of using fine-tuned Small Language Models (SLMs) as a sustainable alternative for predefined tasks. Here, we present a comparative analysis of
the performance-emissions trade-off between LLMs and fine-tuned SLMs across selected tasks under Natural Language Processing, Reasoning and Programming. Our results show that in four out of the six selected tasks, SLMs maintained comparable performances for a significant reduction in carbon emissions during inference.
Our findings demonstrate the viability of smaller models in mitigating the environmental impact of resource-heavy LLMs, thus advancing towards sustainable, green AI.
\end{abstract}


\section{Introduction}
Large Language Models (LLMs), such as OpenAI’s GPT series, Meta AI’s Llama series, and Google's Gemini, have pervaded modern-day academia and become an integral part of the average consumer’s daily life. By September 2024, roughly 18\% of consumer complaints, 24\% of corporate press releases, just under 10\% of job postings in small firms, and nearly 14\% of UN press releases were identified as being assisted by large language models [1].
However, this pervasion comes at a great environmental cost. From energy-intensive training processes to billions of inferences daily, the life cycles of these models generate significant amounts of carbon dioxide (CO$_2$). The final training run for OpenAI's GPT-3 model (with 175 billion parameters), which did not include the significant carbon costs of hardware manufacturing, model development, or inference, emitted an estimated over 550 metric tons of CO$_2$ [2]. To illustrate these "hidden" costs, researchers from the Allen Institute for Artificial Intelligence, in the process of building their own series of smaller models (up to 13 billion parameters), calculated their total footprint. They highlighted that when accounting for the full life cycle, including hardware manufacturing, model development, and final training runs, their models emitted 493 metric tons of carbon. This amount is equivalent to powering approximately 98 US homes for one year [3]. Although training generates a significant portion of the emissions, inference occurs consistently and on a large scale. Recent estimates suggest that inference can take up to 90\% of a model’s total life cycle energy use [4]. This highlights the urgency of extensive inquiry into the matter, prompting research on fine-tuning Small Language Models for certain tasks instead. This paper proposes Small Language Models (SLMs) being almost as good as LLMs at certain tasks, while emitting significantly less CO$_2$, is a worthwhile trade-off.

In this paper, we aim to compare the performance of fine-tuned SLMs and LLMs on a variety of tasks along with calculating the carbon emissions for the same, focusing solely on inference time emissions. We grouped our tasks into three main categories: Language, Reasoning, and Programming. We fine-tuned three SLMs for each task, benchmarking them against open source LLMs such as Qwen 3 (235 billion parameters), Mistral Instruct (7 billion parameters), and DeepSeek R1 (671 billion parameters). We defined an LLM as a Language Model with more than 1 billion parameters. To calculate emissions, we used eco2AI, an open-source library capable of tracking equivalent carbon emissions while training or inferring Python-based AI models accounting for energy consumption of CPU, GPU, RAM devices. It focuses on accuracy of energy consumption tracking and correct regional CO$_2$2 emissions accounting due to precise measurement of process loading, extensive database of regional emission coefficients and CPU devices [5].
Through our research, we empirically investigate whether fine-tuned SLMs ($\leq 250$ million parameters) can achieve performance parity with LLMs ($\leq 1$ billion parameters) across specific tasks, while cutting down inference emissions by orders of magnitude, addressing a critical gap in sustainable AI research.

Additionally, we highlight the need for consistent measures of the environmental impact of LLMs in order to make informed sustainable policy decisions. Estimates of per-prompt energy consumption for similar AI tasks can vary by an order of magnitude due to the lack of consensus on the measurement boundary. Environment metrics should be standardized and comprehensive to allow meaningful comparison across different models and providers [6].
 
\section{Related Works}
Although SLMs deliver solid results on many NLP benchmarks, they generally fall short of large models in terms of overall accuracy and generalization capabilities. “There is a non-negligible body of empirical evidence demonstrating the superiority of large language models in general language understanding over small models of the same generation” [7]. This is consistent with language model scaling laws, first formalized by Kaplan et al. (2020) [8]. These laws empirically demonstrate that model performance (measured by loss) improves as a predictable power-law function of the model's parameter count, the size of the training dataset, and the total compute used for training. This enables better performance across a variety of specialized tasks, including text generation, translation, and reasoning. However, it is possible to improve the accuracy and performance of these small models through techniques such as fine-tuning [9].

Previously, Kumar et al. [10] performed a study that demonstrated that fine-tuning smaller SLMs and VLMs, models ranging up to 7 billion parameters, could achieve performances comparable to larger models for tasks such as Image Captioning, Visual Question Answering, Dialogue Summarization, and Text-to-SQL conversion, with significantly lower environmental impact. Extending prior efforts like Kumar et al., which examined up to 7 billion parameters for multimodal tasks, we evaluate ultra-compact SLMs ($\leq 250$ million parameters) across an expanded set of unimodal tasks. The focus on smaller scales probes enhanced accessibility and reduced reliance on emission-intensive LLMs. Unlike most existing studies, which prioritize training emissions, our analysis centers on inference, the dominant life cycle phase [4]. 
To our knowledge, this is the first systematic comparison of such compact SLMs against LLMs for inference emissions and performance in diverse unimodal domains.

Prior studies predominantly focused on the training phase of models, giving secondary focus to the inference phase despite its significant cumulative energy footprint over a model's lifespan. Research shows that response characteristics such as response token length and response duration and total inference duration are strongly correlated with energy usage, with coefficients of 0.846, 0.625, and 0.618 respectively, suggesting that managing response generation is key to energy savings. Prompt complexity seems to not affect energy use significantly [11]. However, there is a lack of first-party data on environmental metrics from the largest model providers [6], making it difficult to assess the true environmental cost of full deployment and hindering efforts to develop effective reduction strategies.

Regulatory acts, such as the European Union Artificial Intelligence Act mandate reporting energy consumption only during the model's development phase. The AI Act's environmental provisions were diluted during negotiations, leading to critical omissions including the failure to mandate Sustainability Impact Assessments for all AI systems. Furthermore, the EU data center regulation remains incomplete, lacking binding efficiency and renewable energy targets for data centers [12].

\section{Methodology and Results}

We conducted our analysis by first selecting appropriate tasks and suitable SLMs for fine-tuning. The tasks were organized into three broad categories: Natural Language Processing (NLP), Reasoning, and Programming. For each category, we chose two representative subtasks: for NLP, Sentiment Analysis and Content Creation; for Reasoning, Chain of Thought Reasoning and Natural Language Inference; and for Programming, Code Summarization and Code Generation.

We then benchmarked the performance of our models against open-source LLMs: Mistral 7B, Qwen 3 235B A22B, and DeepSeek R1 671B and calculated the performance-emissions trade-off using the \texttt{eco2ai} library. We limited the SLM parameters to $\leq 250$ million and LLM parameters to  $\geq 1$ billion for this study.

Given the absence of public inference emission data for our benchmark LLMs, we derived estimates by scaling from GPT-3’s estimated 4.32 g CO$_2$ per query, assumed a query on average is 100 tokens, adjusted linearly for active parameters. This approach, while approximate, aligns with established methodologies [4,13], and accounts for hardware and geographic variances via \texttt{eco2AI}.

Carbon per token per parameter:
\begin{equation} \label{eq:k_constant}
\begin{split}
  K &= \frac{4.32 \text{ grams} / 100 \text{ tokens}}{175 \text{ billion parameters}} \\
  K &= 0.0002469 \text{ g} / (\text{billion-param} \cdot \text{token})
\end{split}
\end{equation}

Carbon per token for each model:
\begin{equation} \label{eq:carbon_token}
  \text{Carbon}_{\text{token}} = K \times P_{\text{active}}
\end{equation}

Using the above formulas, we can now derive the following (where $P_{\text{active}}$ is the number of active parameters in billions):

\begin{itemize}
  \item DeepSeek-R1-0528 ($P_{\text{active}}=37$): \\
        $0.0002469 \times 37 = 0.00914 \text{ g/token}$
        
  \item Qwen3-235B-A22B ($P_{\text{active}}=22$): \\
        $0.0002469 \times 22 = 0.00543 \text{ g/token}$
        
  \item Mistral 7B ($P_{\text{active}}=7.3$): \\
        $0.0002469 \times 7.3 = 0.00180 \text{ g/token}$
\end{itemize}

Carbon per inference (g CO$_2$e) = $\text{Carbon}_{\text{token}} \times \text{Tokens}_{\text{task}}$
\begin{equation} \label{eq:carbon_inference}
  \text{Carbon}_{\text{inference}} = \text{Carbon}_{\text{token}} \times \text{Tokens}_{\text{task}}
\end{equation}

DeepSeek-R1-0528 utilizes a Mixture-of-Experts architecture with 37 billion active parameters from 671 billion total parameters [14], while Qwen3-235B-A22B employs 22 billion active parameters from 235 billion total [15]. Mistral 7B Instruct operates as a dense model with 7.3 billion parameters [16]. Token consumption estimates for each task are: Sentiment Analysis (150 tokens), Content Creation (800 tokens), Chain-of-Thought Reasoning (600 tokens), Natural Language Inference (200 tokens), Code Summarization (400 tokens), and Code Generation (400 tokens). We calculated emissions per query using the above estimations.

\subsection{Natural Language Processing (NLP)}

\subsubsection{Sentiment Analysis (Binary)}
\begin{itemize}
    \item \textbf{Models:} ELECTRA (14 million parameters), DistilBERT (66 million parameters), DistilRoBERTa-base (82 million parameters)
    \item \textbf{Dataset:} Yelp Polarity Reviews [17]
    \item \textbf{Primary Evaluation Metric:} Accuracy
\end{itemize}

For this task, we performed binary sentiment analysis on a dataset of Yelp reviews. Even the smallest model (ELECTRA, 14 million parameters) demonstrated improved accuracy compared to the LLMs. The inference emissions for the fine-tuned models were low, ranging from 60 milligrams for the larger model to 0.1 milligrams per query for the smaller one. Performance differences between these models were negligible.
\begin{table}[H]
\centering
\caption{Sentiment Analysis Results Comparison}
\renewcommand{\arraystretch}{1.5}
\begin{tabular}{|c|c|c|}
\hline
\textbf{Model} & \textbf{Accuracy} & \textbf{Inference Emissions (g/query)} \\
\hline
ELECTRA (14M) & 0.95 & 0.0001 \\
\hline
DistilBERT (66M) & 0.95 & 0.0008 \\
\hline
DistilRoBERTa-base (82M) & 0.96 & 0.0696 \\
\hline
Qwen3 A22B (235B) & 0.94 & 0.82 \\
\hline
Mistral-7B-Instruct-v0.2 & 0.68 & 0.27 \\
\hline
DeepSeek R1 0528 (671B) & 0.84 & 1.37 \\
\hline
\end{tabular}
\renewcommand{\arraystretch}{1.0}
\end{table}

\subsubsection{Content Creation (Short Story Writing)}

\begin{itemize}
    \item \textbf{Models:} DistilGPT2 (82M), GPT-2 (137M), Pythia (160M)
    \item \textbf{Dataset:} WritingPrompts [18]
    \item \textbf{Primary Evaluation Metrics:} Perplexity, BERT Score F1
\end{itemize}

For this task, we generated short stories based on a single-line prompt. We observed consistent BERT Score F1 and Perplexity metrics across the fine-tuned SLMs. The inference emissions for the models were very low, ranging from 6 to 10 milligrams per query.

\begin{table}[H]
\caption{Content Creation Results Comparison}
\renewcommand{\arraystretch}{1.5}
\begin{tabular}{|c|c|c|c|}
\hline
\textbf{Model} & \textbf{\makecell{BERT Score\\ F1}} & \textbf{Perplexity} & \textbf{\makecell{Inference \\ Emissions\\(g/query)}} \\
\hline
GPT-2(137M) & 0.79 & 24.14 & 0.0063\\
\hline
DistilGPT2 (82M) & 0.79 & 30.27 & 0.0063 \\
\hline
Pythia (160M) & 0.70 & 34.35 & 0.0105 \\
\hline
Qwen3 A22B (235B) & 0.81 & 58.86 & 4.35 \\
\hline
Mistral-7B-Instruct-v0.2 & 0.81 & 19.08 & 1.44 \\
\hline
DeepSeek R1 0528 (671B) & 0.80 & 70.01 & 7.31 \\
\hline
\end{tabular}
\renewcommand{\arraystretch}{0}
\end{table}

\subsection{Reasoning}
\subsubsection{Natural Language Inference (NLI)}
\begin{itemize}
\item \textbf{Models:} ELECTRA (14M), FLAN-T5 Base (250M), deBERTa v3 Base (184M)
\item \textbf{Dataset:} Stanford NLP/SNLI [19]
\item \textbf{Primary Evaluation Metric:} Accuracy
\end{itemize}

For this task, fine-tuned models of FLAN-T5 Base (250 million parameters) and deBERTa v3 Base (184 million parameters) were able to outperform all of the LLMs, achieving accuracies of 88.0\% and 88.39\%, respectively. The highest accuracy among the LLMs was 80.0\%, achieved by DeepSeek-R1 0528 (671 billion parameters). The inference emissions for the fine-tuned models were very low, around 1 milligram per query for all models, despite their varying performances.

\begin{table}[H]
\centering
\caption{Natural Language Inference Results Comparison}
\renewcommand{\arraystretch}{1.5}
\begin{tabular}{|l|c|c|}
\hline
\textbf{Model} & \textbf{Accuracy} & \textbf{\makecell{Inference \\ Emissions (g/query)}} \\
\hline
FlanT5-base (250M) & 0.88 & 0.0019 \\
\hline
ELECTRA (14M) & 0.36 & 0.0015 \\
\hline
DeBERTa v3 base (184M) & 0.88 & 0.0014 \\
\hline
Qwen2.5-BA22B (235B) & 0.56 & 1.09 \\
\hline
Mistral-7B-Instruct-v0.2 & 0.62 & 0.36 \\
\hline
DeepSeek-R1-0528 (671B) & 0.80 & 1.83 \\
\hline
\end{tabular}
\renewcommand{\arraystretch}{1}
\end{table}

\subsubsection{Chain-of-Thought Reasoning}

\begin{itemize}
\item \textbf{Models:} T5-Reasoning, FLAN-T5 Small (80M), FLAN-T5 Base (250M)
\item \textbf{Dataset:} The CoT Collection [20]
\item \textbf{Primary Evaluation Metric:} Modified G-Eval
\end{itemize}

The accuracy of fine-tuned SLMs for task were quite low. Prior literature indicates that advanced reasoning typically emerges in models exceeding tens of billions of parameters [21]. Therefore, it was expected that these models struggle with complex reasoning. The inference emissions for the fine-tuned models were very low, ranging from 0.5 milligrams to 1 milligram per query.

The evaluation method is described as follows:
G-Eval is a framework that uses Large Language Models with Chain-of-Thought (CoT) reasoning and a form-filling paradigm to assess the quality of Natural Language Generation (NLG) outputs [22]. We modified G-Eval and prompted OpenAI’s GPT-4.1 Nano as an evaluator to measure the accuracy of reasoning steps and final answer accuracy as separate metrics (as given in the Appendix).

Based on this we carried out our evaluation.

\begin{table}[H]
\centering
\caption{CoT Reasoning Results Comparison}
\renewcommand{\arraystretch}{1.5}
\begin{tabular}{|c|p{1.25cm}|p{1.25cm}|c|}
\hline
\textbf{Model} & \textbf{Reasoning Accuracy (Modified G-Eval)} & \textbf{Answer Accuracy (Modified G-Eval)} & \textbf{\makecell{Inference \\ Emissions \\(g/query)}} \\
\hline
t5-reasoning & 19.4 & 31.4 & 0.0010\\
\hline
FLANt5-small (80M) & 30.2 & 31.4 & 0.0016\\
\hline
FLANt5-base (250M) & 47.0 & 46.8 & 0.0005\\
\hline
Qwen3 A22B (235B) & 54 & 57 & 3.26\\
\hline
Mistral-7B-Instruct-v0.2 & 45 & 47 & 1.08\\
\hline
DeepSeek R1 0528 (671B) & 64 & 68.7 & 5.48\\
\hline
\end{tabular}
\renewcommand{\arraystretch}{1}
\end{table}

\subsection{Programming}
\subsubsection{Code Summarization (Python)}
\begin{itemize}
\item \textbf{Models:} CodeT5-Small (60M parameters), CodeParrot-Small (110M parameters), TinyCodeLM (150M parameters)
\item \textbf{Dataset:} CodeSearchNet [23]
\item \textbf{Primary Evaluation Metrics:} BLEU Score, ROUGE-L Score
\end{itemize}

In this task, Qwen3-235B-A22B (235 billion parameters) significantly underperformed the SLMs, achieving a BLEU score of 4.57 and a ROUGE-L score of 0.18. Both Mistral-7B-Instruct-v0.2 and DeepSeek-R1 0528 (671 billion parameters) surpassed CodeT5-Small but trailed behind CodeParrot-Small and TinyCodeLM. The inference emissions for the fine-tuned SLMs ranged from 90 milligrams per query for the larger models to 0.5 milligrams per query for the relatively smaller ones. Despite these differences, the overall performance variation among the SLMs was negligible.

\begin{table}[H]
\centering
\caption{Code Summarization Results Comparison}
\renewcommand{\arraystretch}{1.5}
\begin{tabular}{|c|c|c|c|}
\hline
\textbf{Model} & \textbf{\makecell{BLEU \\Score}} & \textbf{\makecell{ROUGE-L \\Score}} & \textbf{\makecell{Inference \\ Emissions \\(g/query)}} \\
\hline
CodeT5-small (60M) & 3.83 & 0.68 & 0.0056 \\
\hline
CodeParrot-small (110M) & 25.60 & 0.43 & 0.0055 \\
\hline
TinyCodeLM (150M) & 24.21 & 0.41 & 0.0973 \\
\hline
Qwen2.5-BA22B (235B) & 4.57 & 0.18 & 2.17 \\
\hline
Mistral-7B-Instruct-v0.2 & 18.77 & 0.33 & 0.72 \\
\hline
DeepSeek-R1-0528 (671B) & 11.27 & 0.24 & 3.66 \\
\hline
\end{tabular}
\renewcommand{\arraystretch}{1}
\end{table}
\subsubsection{Code Generation (Python)}
\begin{itemize}
\item \textbf{Models:} CodeT5-Small (60M parameters), CodeParrot-Small (110M parameters), TinyCodeLM (150M parameters)
\item \textbf{Dataset:} HumanEval [24]
\item \textbf{Primary Evaluation Metrics:} pass@1 Score
\end{itemize}
For the code generation task, we evaluated the models' ability to generate functional Python code from docstrings. The results align with our hypothesis and existing literature, which suggest that complex reasoning and generation tasks require model scale.

The fine-tuned SLMs, while optimized for code, showed very low performance, with the best model, TinyCodeLM (150M), achieving only a 12.8\% pass@1 score. The larger CodeT5+ (220M) model performed similarly at 12.0\%. In stark contrast, the LLMs demonstrated significantly stronger capabilities. Mistral-7B-Instruct achieved a 36.0\% pass@1 score, and the Qwen3 model scored 29.8\%. The DeepSeek R1 model, a top-tier reasoning model, showed the highest performance at 65.9\%.

\begin{table}[H]
\centering
\caption{Code Generation Results Comparison}
\renewcommand{\arraystretch}{1.5}
\begin{tabular}{|c|c|c|c|}
\hline
\textbf{Model} & \textbf{pass@1 Score} & \textbf{\makecell{Inference \\ Emissions (g/query)}} \\
\hline
CodeT5-small (60M) & 0.120 & 0.0056 \\
\hline
CodeParrot-small (110M) & 0.038 & 0.0055\\
\hline
TinyCodeLM (150M) & 0.128 & 0.0973  \\
\hline
Qwen2.5-BA22B (235B) & 0.298 & 2.1700  \\
\hline
Mistral-7B-Instruct-v0.2 & 0.360 & 0.7200 \\
\hline
DeepSeek-R1-0528 (671B) & 0.659 & 3.6600 \\
\hline
\end{tabular}
\renewcommand{\arraystretch}{1}
\end{table}

Therefore, we can observe that comparable performance between LLMs and fine-tuned LLMs was achieved in four out of six tasks, namely, Sentiment Analysis and Content Creation (under NLP), Natural Language Inference (under Reasoning), and Code Summarization (under Programming). There was a tremendous reduction in inference emissions for all tasks.

\section{Limitations}

Although our research says that fine-tuned SLMs can replace LLMs for certain tasks, future work should delineate task boundaries more precisely through expanded benchmarks. Another limitation is that we could only evaluate open-source LLMs due to resource constraints, but further research should benchmark against more advanced models such as GPT-5. While this study focuses on fine-tuning to improve SLM performance, there are emerging alternative methods available as well. Mixture-of-Experts (MoE) architectures, dynamic model routing, knowledge distillation, parameter-efficient fine-tuning methods such as LoRA are able to achieve improved performance with reduced compute. There is potential for further research with the utilization of these methods. 

Finally, since our experiments were performed on a very small scale and focused on Small Language Models $\leq 250$ million parameters, further research is needed to determine the “sweet spot” in model size where SLMs can deliver the optimal balance between performance and emissions relative to LLMs.

\section{Conclusion}

This study confronts what we are calling the \textit{LLM Energy Dilemma}. By demonstrating SLM parity with a dramatic cutdown in inference emissions across 4 out of 6 tasks, we provide actionable evidence for shifting towards energy efficient AI architectures. 

Significant improvements were observed in performance after fine-tuning for Sentiment Analysis, Content Creation, Natural Language Inference, and Code Summarization. The CO$_2$ emitted was up to 13,000 times, 1,100 times, 1,300 times and 660 times lesser per query respectively. The performance for Content Creation remained largely unchanged, but the emissions were substantially lower, up to 1,200 times per query.

Despite fine-tuning, SLMs were unable to match LLMs for Chain of Thought Reasoning and Code Generation. This was due to the models having fewer than a billion parameters, which are recognized as inadequate for such complex tasks.

When it comes to the question of the LLM energy dilemma, and whether it is worth it to fine-tune SLMs for certain tasks – the answer seems to be \textit{sometimes}. While SLMs may not match LLMs on all tasks, there is significant potential to capitalize on the tasks where they excel. This points towards a future of more energy-efficient AI deployment, where curated task-specific fine-tuning can achieve strong performance with a drastically lower carbon footprint.

Regulatory efforts should include mandates for providers to disclose the energy costs of inferences, as well as an estimated number of inferences on a monthly/yearly level, so that overall inference energy consumption may be calculated. Mandatory Sustainable Impact Assessments for all AI systems so that environmental impact may be taken into consideration should also be included [12].

Firms must balance the benefits of AI adoption with strategies to mitigate the environmental costs associated with AI implementation.

Corporations working on incorporating LLMs into their product workflows must balance the benefits of AI adoption with strategies to mitigate the environmental costs associated with AI implementation. Additionally, the \textit{LLM Energy Dilemma} extends beyond environmental impact for these corporations, giving way to financial implications as well. This can be analyzed using a Total Cost of Ownership (TCO) framework, which includes capital expenditures for hardware and operational expenditures for power and maintenance [26]. Inference is not a one-time capital expense but rather a continuous operational cost, depending on GPU selection and energy consumption [25]. Deploying LLMs requires significant capital expenditure for clusters of expensive, high- performing GPUs. This is coupled with high operational expenditure due to inferences, as our data reflects. Our findings indicate that for specific tasks, using fine-tuned SLMs (requiring more economical hardware) lowers both the capital and operational expenditure, resulting in a significantly lower TCO. This demonstrates that sustainable, Green AI is not just ethically and environmentally beneficial, but also has a financial advantage [26].

\section*{Appendix}
\section*{Modified G-Eval Prompt}

This is the prompt to evaluate Chain-of-Thought outputs \indent as Answering and Reasoning accuracy.\\

\begin{quote}
``You will be given a question and a model-generated response that includes both a reasoning process (Chain-of-Thought) and a final answer. You will also be provided with the correct answer (target) from the dataset. Your task is to evaluate the model's response on two metrics:\\

\begin{enumerate}
\item Whether the final answer is correct.
\item Whether the reasoning (Chain-of-Thought) is valid.\\
\end{enumerate}

Evaluation Criteria:
\begin{enumerate}
    \item {Answer (Correct/Incorrect)}

Mark as Correct if the model's final answer matches or reasonably paraphrases the target answer. Minor variations in phrasing, synonyms, or format are acceptable if the meaning remains the same.

Mark as Incorrect if the final answer contradicts the target or conveys a different meaning, even if the reasoning appears sound.

    \item {Reasoning (Valid/Invalid)}

Mark as Valid if the Chain-of-Thought is logically coherent, relevant to the question, and follows a sensible step-by-step process, even if the final answer is wrong.

Mark as Invalid if the reasoning contains logical flaws, factual errors, irrelevant steps, or faulty deductions, regardless of the final answer.\\
\end{enumerate}

Evaluation Steps:
\begin{enumerate}
    \item Read the question and the dataset's correct answer (target).
    \item Read the model's full response, including the reasoning and final answer.
    \item Judge whether the final answer is Correct or Incorrect based on semantic alignment with the target. Judge whether the reasoning is Valid or Invalid based on logical coherence and relevance.\\
\end{enumerate}

Question: 
{\{Source}\}

Target Answer: 
{\{Target}\}

Model Response: 
{\{Inference}\}\\

Evaluation Form (select ONLY ONE for each). Do not provide an explanation, only your final answer. Give outputs of the form:

Answer: Correct or Incorrect

Reasoning: Valid or Invalid"
\end{quote}
\end{document}